\documentclass[10pt,twocolumn,letterpaper]{article}

\usepackage[pagenumbers]{cvpr} 

\usepackage{graphicx}
\usepackage{amsmath}
\usepackage{amssymb}
\usepackage{booktabs}
\usepackage[accsupp]{axessibility}  

\usepackage[pagebackref,breaklinks,colorlinks]{hyperref}

\usepackage[capitalize]{cleveref}
\crefname{section}{Sec.}{Secs.}
\Crefname{section}{Section}{Sections}
\Crefname{table}{Table}{Tables}
\crefname{table}{Tab.}{Tabs.}

\def\cvprPaperID{****} 
\def\confName{CVPR}
\def\confYear{2023}

\begin{document}

\title{Rethinking Dilated Convolution for Real-time Semantic Segmentation}

\author{Roland Gao\\
University of Toronto\\
roland.gao@mail.utoronto.ca\\
}
\maketitle

\begin{abstract}
The field-of-view is an important metric when designing a model for semantic segmentation. To obtain a large field-of-view, previous approaches generally choose to rapidly downsample the resolution, usually with average poolings or stride 2 convolutions. We take a different approach by using dilated convolutions with large dilation rates throughout the backbone, allowing the backbone to easily tune its field-of-view by adjusting its dilation rates, and show that it's competitive with existing approaches. To effectively use the dilated convolution, we show a simple upper bound on the dilation rate in order to not leave gaps in between the convolutional weights, and design an SE-ResNeXt inspired block structure that uses two parallel $3\times 3$ convolutions with different dilation rates to preserve the local details. Manually tuning the dilation rates for every block can be difficult, so we also introduce a differentiable neural architecture search method that uses gradient descent to optimize the dilation rates. In addition, we propose a lightweight decoder that restores local information better than common alternatives. To demonstrate the effectiveness of our approach, our model RegSeg achieves competitive results on real-time Cityscapes and CamVid datasets. Using a T4 GPU with mixed precision, RegSeg achieves 78.3 mIOU on Cityscapes test set at $37$ FPS, and 80.9 mIOU on CamVid test set at $112$ FPS, both without ImageNet pretraining.
\end{abstract}

\section{Introduction}
\label{sec:intro}

Semantic segmentation is the task of assigning a class to every pixel in the input image. Applications of it include autonomous driving, natural scene understanding, and robotics. It is also the groundwork for the bottom-up approach\cite{panoptic_deeplab_2020} of panoptic segmentation, which, in addition to assigning a class to every pixel, separates instances of the same class.

The field-of-view is an important metric when designing a model for semantic segmentation. To quickly increase the field-of-view, previous advances in semantic segmentation generally adapt a backbone designed for ImageNet\cite{deng2009imagenet} and add a context module with large average poolings like PPM\cite{pspnet} or large dilation rates like ASPP\cite{deeplabv3}. However, the ImageNet backbone, which performs most of the computation, does not have the field-of-view required to encode high resolution images. Some recent advances such as DDRNet23\cite{ddrnet} and STDC\cite{stdc} have already designed backbones specifically for semantic segmentation. Instead of applying rapid down sampling (DDRNet23) or using many more $3\times 3$ convs (STDC) to increase the field-of-view, we keep the features at $1/16$ of the original resolution and use dilated convolutions with large dilation rates throughout the backbone, allowing its field-of-view to be precisely tuned by adjusting its dilation rates. To the best of our knowledge, we are the first to use large dilation rates throughout the backbone while achieving competitive results on standard semantic segmentation benchmarks. 

\begin{figure}
  \centering
    \includegraphics[width=1.0\linewidth]{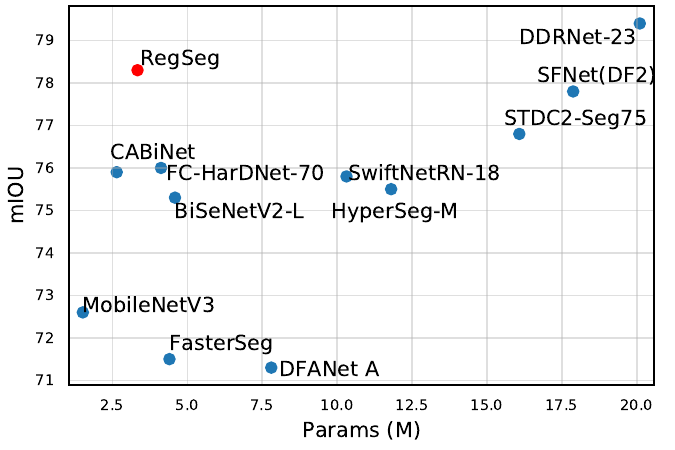}

   \caption{Params vs mIOU on Cityscapes test set. Our model is in red, while other models are in blue. We achieve SOTA params-accuracy trade-off.}
   \label{fig:miouVsParams}
\end{figure}

\begin{figure}
  \centering
    \includegraphics[width=0.8\linewidth]{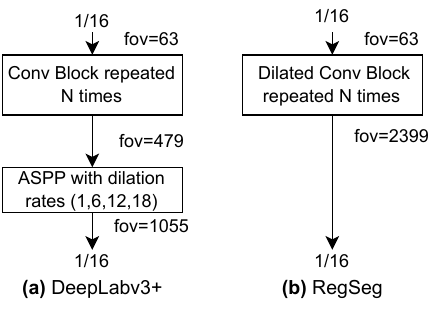}

   \caption{RegSeg vs DeepLabv3+ architecture design. RegSeg's simple design achieves larger field-of-view than DeepLabv3+ with RegNetY-600MF backbone (2399 vs 1055). DeepLabv3+ performs most of its computation in the conv blocks, where it has a low field-of-view. RegSeg performs most of its computation in the dilated blocks, where the field-of-view is much larger.}
   \label{fig:regseg_vs_deeplab}
\end{figure}

To effectively use the dilated convolution, we have to solve many problems that previous researchers have struggled with. First, we show a simple upper bound on the dilation rate in order to not leave gaps in between the convolutional weights. Next, to preserve the local details, we modify the SE-ResNeXt\cite{hu2018squeeze,resnext} block such that, when applying the $3\times 3$ group conv, we use a small dilation rate for half of the groups and another larger dilation rate for the other half. We adapt the fast RegNetY-600MF\cite{regnet,regnetz} for semantic segmentation by swapping out the original SE-ResNeXt block, which they call the Y block, with our modified dilated block, which we call the D block. As shown in \cref{fig:regseg_vs_deeplab}, RegSeg's simple design achieves higher field-of-view than the standard DeepLabv3+ (2399 vs 1055) without using ASPP or PPM. While DeepLabv3+ performs most of its computations in a low field-of-view ranging from 63 to 479, RegSeg performs most of its computations in a much larger field-of-view ranging from 63 to 2399.

Manually tuning the dilation rates for every block can be difficult, so we also introduce a differentiable neural architecture search method that uses gradient descent to optimize the dilation rates. We accomplish this by viewing dilated conv as a special type of deformable conv\cite{dai2017deformable} and using a trainable parameter $r$ to generate the offsets required for deformable conv. Deformable conv's implementation allows back propagation on the offsets, so we use it to optimize the dilation rate $r$.

We also propose a lightweight decoder that effectively restores the local details lost in the backbone. Previous decoders such as the one in DeepLabv3+\cite{deeplabv3plus2018} are too slow to run in real-time, and common lightweight alternatives such as LRASPP\cite{mobilenetv32019} are not as effective. Our decoder is 1.0\% better than LRASPP under the same training setting.

RegSeg's simple design allows it to be extremely efficient, achieving state-of-the-art params-accuracy trade-off and FLOPS-accuracy trade-off on Cityscapes\cite{cityscapes}, as shown in \cref{fig:miouVsParams} and \cref{tab:cityscapesComparison}. RegSeg achieves those results without pretraining on ImageNet\cite{deng2009imagenet}, thus being data-efficient as well. RegSeg demonstrates competitive runtime performance as well when timed on Nvidia's T4 GPU and Mac's M1 GPU, as shown in \cref{tab:cityscapesComparison}.


\section{Related works}
\subsection{Network design}
The models found on ImageNet play an important role in general network design, and their improvements often transfer to other domains such as semantic segmentation. RegNet\cite{regnet,regnetz} finds many improvements to the ResNeXt\cite{resnext} architecture by using random search to run numerous experiments and analyzing trends to reduce the search space. They provide models across a wide range of flop regimes, and the models outperform EfficientNet\cite{tan2019efficientnet} under comparable training settings. EfficientNetV2\cite{efficientnetv2} is the improved version of EfficientNet and trains faster by using regular convs instead of depthwise convs at the higher resolutions.

\subsection{Semantic segmentation}

Fully Convolutional Networks (FCNs)\cite{overfeat,fcn} are shown to beat traditional approaches in the task of segmentation. DeepLabv3\cite{deeplabv3} uses dilated conv in the ImageNet pretrained backbone to reduce the output stride to 16 or 8 instead of the usual 32, and increases the receptive field by proposing the Atrous Spatial Pyramid Pooling module (ASPP), which applies parallel branches of convolutional layers with different dilation rates. PSPNet\cite{pspnet} proposes the Pyramid Pooling Module (PPM), which applies parallel branches of convolutional layers with different input resolutions by first applying average poolings.

DeepLabv3+\cite{deeplabv3plus2018} builds on top of DeepLabv3 by adding a simple decoder with two $3\times 3$ convs at output stride $4$ to improve the segmentation quality around boundaries. HRNetV2\cite{hrnetv2} keeps parallel branches with different resolutions right in the backbone, with the finest one at output stride $4$. 

In our paper, we adapt RegNet for semantic segmentation by directly using dilated conv with large dilation rates in the backbone and show that it performs better than using DeepLabV3+'s approach of using attaching an ASPP.


\subsection{Real-time semantic segmentation}
MobilenetV3 uses the lightweight decoder LRASPP\cite{mobilenetv32019} to adapt the fast ImageNet model for semantic segmentation. BiSeNetV1\cite{bisenetv1} and BiSeNetV2\cite{bisenetv2} have two branches in the backbone (Spatial Path and Context Path) and merge them at the end to achieve good accuracy and performance without ImageNet pretraining. SFNet\cite{sfnet} proposes the Flow Alignment Module (FAM) to upsample low resolution features better than bilinear interpolation. STDC\cite{stdc} rethinks the BiSeNet architecture by removing the Spatial Path and designing a better backbone. HarDNet\cite{hardnet} reduces GPU memory traffic consumption by using mostly $3\times 3$ convs and barely any $1\times 1$ convs. DDRNet-23\cite{ddrnet} uses two branches with multiple bilateral fusions between them and appends a new context module called the Deep Aggregation Pyramid Pooling Module (DAPPM) at the end of the backbone.

\section{Methods}

\subsection{Field-of-view}
\label{sec:fov}
\begin{figure}
  \centering
    \includegraphics[width=0.8\linewidth]{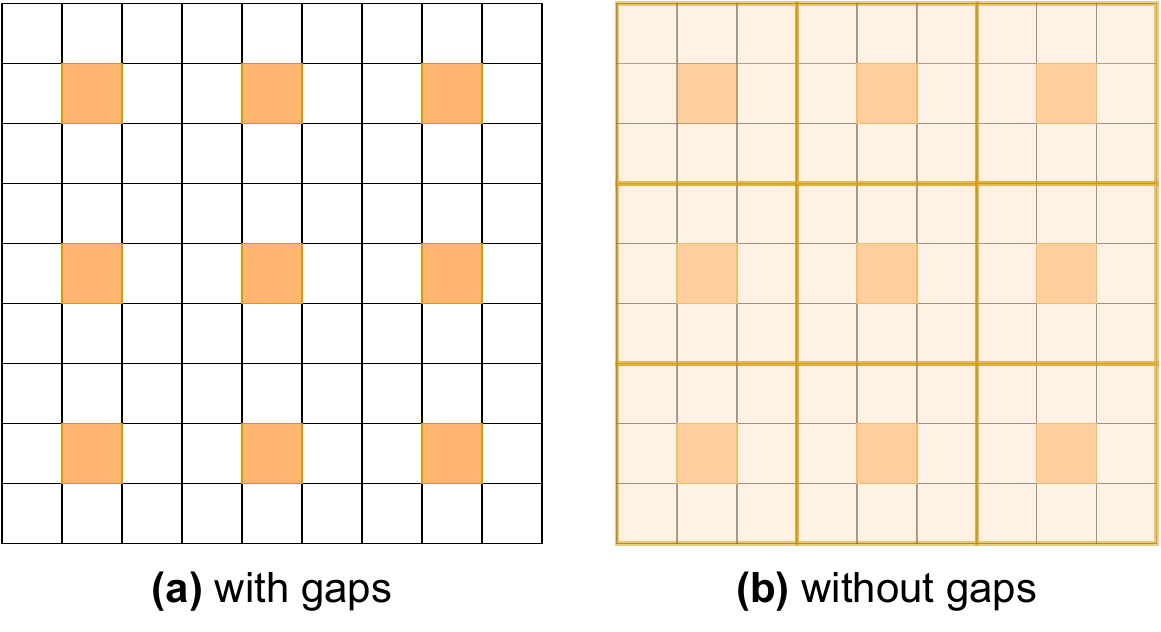}

   \caption{Upper bound $r\leq k/s$ in action. Left: a $3\times 3$ dilated conv with dilation rate 3 leaves gaps in between the weights. Right: Using a $3\times 3$ conv with stride 1 before the dilated conv results in no gaps because the upper bound is satisfied with $r=3,k=3,s=1$.}
   \label{fig:upperbound}
\end{figure}

We are interested in the field-of-view (FOV), also known as the receptive field, of our model gained through convolutions. For example, a composition of two $3\times 3$ convs is equal in kernel size and stride to a $5\times 5$ conv, and we simply say that the field-of-view is $5$. More generally, the field-of-view of a composition of convs can be calculated iteratively as described in FCN\cite{fcn}. Suppose the composition of convs up to the current point is equal in kernel size and stride to one $k\times k$ conv with stride $s$, and we compose it with a $k'\times k'$ conv with stride $s'$. We update $k$ and $s$ by
\begin{align}
k&\leftarrow k+(k'-1)*s\\
s&\leftarrow s*s'
\end{align}
The field-of-view is the final value of $k$.

A $3\times 3$ conv with dilation rate $r$ is equal in field-of-view to a conv with kernel size $2r+1$. However, this field-of-view is only valid if there are no gaps in between the convolution weights, which happens when the following inequality is satisfied.
\begin{align}
r\leq k/s
\end{align}
\cref{fig:upperbound} gives a simple example of this upper bound in action. Using a $3\times 3$ conv with dilation rate $3$ as the first operation leaves gaps in between weights, as shown in \cref{fig:upperbound}a. If we first use a $3\times 3$ conv with stride 1 before the dilated conv, the inequality is satisfied, so there will not be any gaps, as shown in \cref{fig:upperbound}b.

If the composition of convs before the current point is equal in kernel size and stride to a $k\times k$ conv with stride $s$, then each weight location of the dilated conv corresponds to a $k\times k$ region in the input image. Moving from one weight location to its neighbouring weight location requires shifting $r$ pixels in one direction, which corresponds to shifting the $k\times k$ region in the input image by $sr$ pixels. To not leave any gaps in between the two regions, we need $sr\leq k$, or $r\leq k/s$, as desired. In practice, we choose dilation rates much lower than the upper bounds.

Intuitively, the model should have a large enough field-of-view so that each pixel in the output can see the entire image. For example, if we use the ResNet\cite{resnet} architecture on ImageNet with a testing crop size of $224\times 224$ and look at the feature maps right before the global average pooling, the model needs a field-of-view of at least $224*2-1=447$ for the top-left pixel to see the entire image. Similarly, on Cityscapes with image size $1024\times 2048$, the model needs a field-of-view of 2047 for the top-left pixel of the output to see the bottom-left pixel of the input image, and a field-of-view of 4095 to see the bottom-right pixel of the input image. 


\subsection{Dilated block}
\begin{figure}
  \centering
    \includegraphics[width=1.0\linewidth]{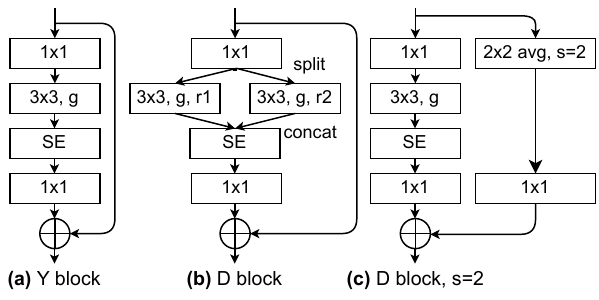}

   \caption{Y block and D block. When $r1=r2=1$, the D block is the same as the Y block.}
   \label{fig:D block}
\end{figure}


Our dilated block (D block) takes inspiration from the Y block of RegNet\cite{regnet}, also known as the SE-ResNeXt block\cite{hu2018squeeze}. The Y block and our new D block utilize group convolutions. Suppose input channels $=$ output channels $=w$, which is always true for the $3\times 3$ convs in the Y block and the D block. A group conv has an attribute called the group width $g$, and $g$ must divide $w$. During the forward pass, the input with $w$ channels are split into $w/g$ groups with $g$ channels each, and a regular conv is applied to each group, and the outputs are concatenated together to form the $w$ channels again.

Since there is a conv for each group, we can apply different dilation rates to different groups to extract multi-scale features. For example, we can apply dilation rate $1$ to half of the groups, and dilation rate $10$ to the other half. This is the key to our D block. \cref{fig:D block}a shows the Y block. \cref{fig:D block}b shows our D block. When $r1=r2=1$, the D block is equivalent to the Y block. In \cref{sec:backboneAblation}, we experiment with some D blocks that have 4 branches of different dilation rates, but find that they are no better than D blocks that have 2 branches. \cref{fig:D block}c shows the D block when stride $=2$. Similar to the ResNet D-variant\cite{resnetd}, we apply a $2\times 2$ average pooling on the shortcut branch when the block's stride $=2$. BatchNorm\cite{batchnorm} and ReLU immediately follow each conv, except that the ReLUs right before the summation are replaced with one after the summation. We use an SE\cite{hu2018squeeze} reduction ratio of $1/4$.


Other multi-path block structures with attention mechanisms have been proposed before, such as SKNet\cite{sknet}, ResNeSt\cite{resnest}, DetectoRS\cite{detectoRS}. Unlike the previous approaches, our main focus is to bring dilated convolution with large dilation rates into the backbone to efficiently increase its field-of-view. For example, ResNeSt does not use dilated convolution at all; SKNet and detectoRS use only small dilation rates that are at most 3. Moreover, the D block is a direct replacement of the Y block in the original RegNet model, with exactly the same number of parameters and FLOPs; other methods use more parameters and FLOPs than the original model to implement the multi-path block structure.



\subsection{Backbone}
\label{sec:backbone}
\begin{table}
  \centering
  \begin{tabular}{c|c|c|c}
    \toprule
    Operator & Stride & \#Channels & \#Repeat \\
    \midrule
    3x3 conv & 2 & 32 & 1\\
    D block & 2 & 48 & 1\\
    D block & 2 & 128 & 3\\
    D block & 2 & 256 & 13\\
    D block & 1 & 320 & 1\\
    \bottomrule
  \end{tabular}
  \caption{Backbone. \#Channels is the number of output channels, and the number of input channels is inferred from the previous block. When stride $=2$ and \#repeat $>1$, the first block has stride 2 and the rest have stride 1.}
  \label{tab:backbone}
\end{table}
\begin{figure}
  \centering
    \includegraphics[width=0.8\linewidth]{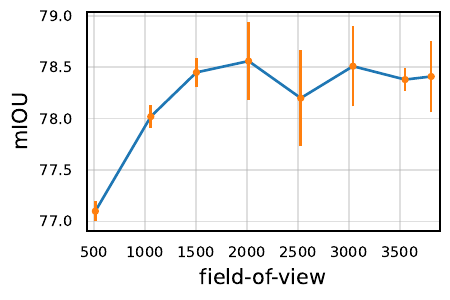}

   \caption{mIOU vs field-of-view on Cityscapes with error bars. The accuracy increases as the field-of-view increases and it stabilizes when the field-of-view reaches around 2000.}
   \label{fig:miou_vs_fov}
\end{figure}
This backbone is directly inspired by RegNetY-600MF\cite{regnet} and has similar params, FLOPs, and runtime. The backbone starts with one $32$-channel $3\times 3$ conv with stride $2$. Then it has one $48$-channel D block at $1/4$ resolution, three 128-channel D block at $1/8$, thirteen 256-channel D block at $1/16$, ending with one $320$-channel D block at $1/16$. Group width $g=16$ for all D blocks. We do not downsample to $1/32$. In a format similar to EfficientNetV2\cite{efficientnetv2}, we display the backbone of RegSeg in \cref{tab:backbone}.

We tune the dilation rates for the last 13 stride 1 blocks while fixing the dilation rate to 1 for all the earlier blocks. We can easily adjust the field-of-view of the backbone by specifying the dilation rates. For example, we can perform an mIOU vs field-of-view analysis by training models of field-of-view from 500 to 3500 in steps of 500, as shown in \cref{fig:miou_vs_fov}. In order to understand what dilation rates would be good for the model, we develop a differentiable neural architecture search method, as discussed in the next section. 

\subsection{Gradient descent on the dilation rates}
\label{sec:dnas}

\begin{figure*}
  \centering
    \includegraphics[width=1.0\linewidth]{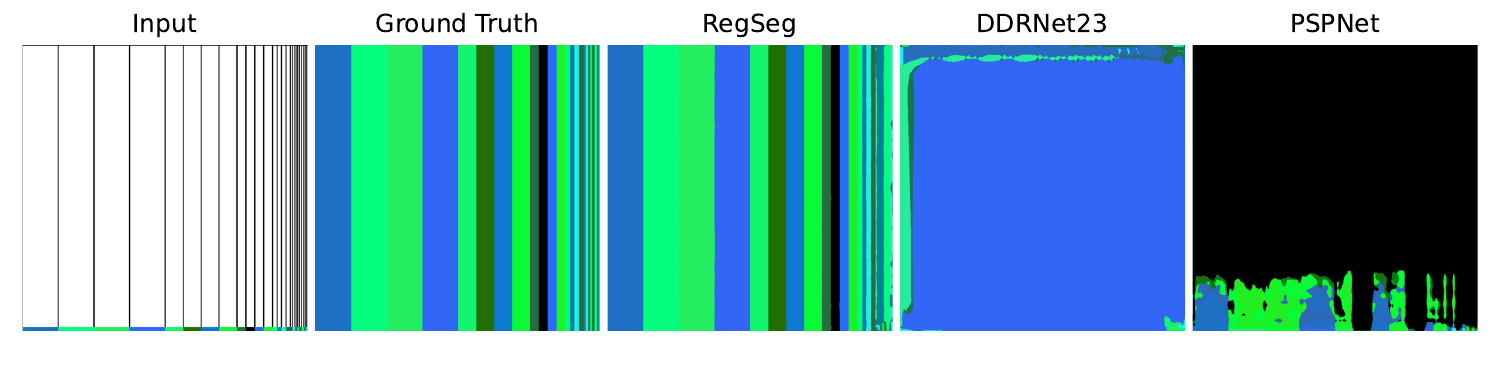}
   \caption{Comparison between RegSeg and DDRNet23 on the toy dataset.}
   \label{fig:toy_comparison}
\end{figure*}

\begin{figure}
  \centering
    \includegraphics[width=0.8\linewidth]{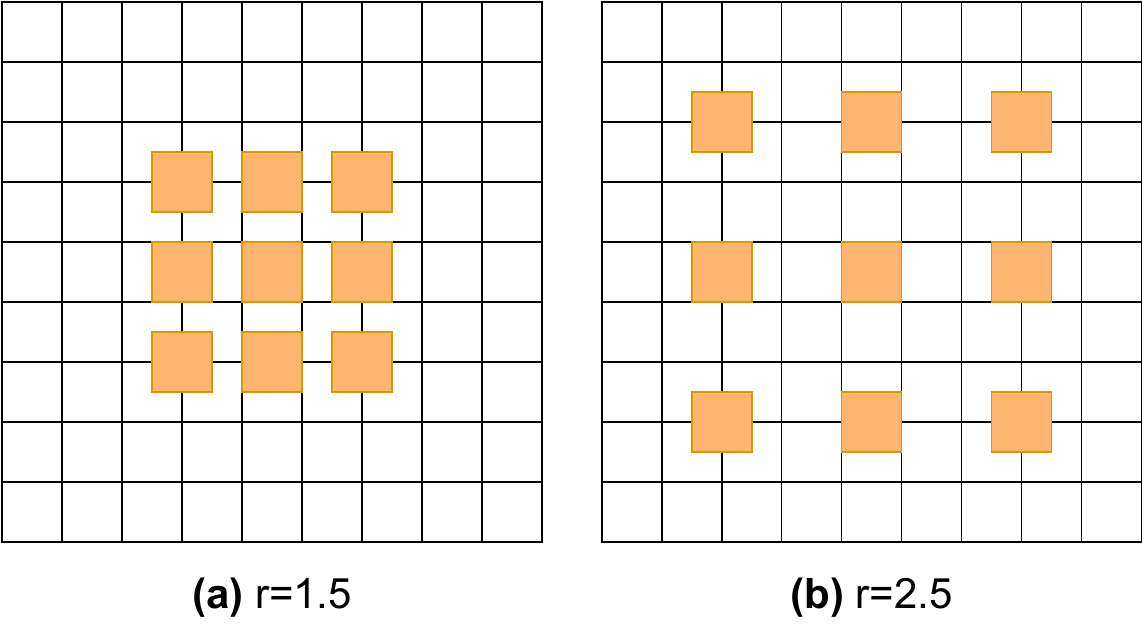}

   \caption{Trainable dilation rates. In order to perform gradient descent on the dilation rates, we extend the type to floating-point. Left: $r=1.5$, right: $r=2.5$.}
   \label{fig:noninteger_rate}
\end{figure}

Manually tuning the dilation rates for every block can be time-consuming, so we introduce a differentiable neural architecture search method that uses gradient descent to optimize the dilation rates. First, we note that dilated conv is a special case of deformable conv\cite{dai2017deformable}, which uses offsets, generally produced by an auxiliary conv, to guide the positions of the weights of the main conv. Even though deformable conv is theoretically more powerful than the dilated conv, it is significantly slower. Besides, its ability to gain the field-of-view is equal to that of the dilated conv because they are both bounded above by $r\leq k/s$, as discussed in \cref{sec:fov}.

For each dilation rate that we want to tune, we introduce a trainable parameter $r$ that represents the dilation rate. We relax the supposedly integer variable to a floating-point and use it to generate the offsets of the deformable conv, such that when $r$ is an integer, the deformable conv does exactly what the dilated conv with dilation rate $r$ does. \cref{fig:noninteger_rate} shows examples of when $r=1.5$ and $r=2.5$. After training the model with deformable conv, we have to convert the model's fractional dilation rates to integers and train the converted model again. For each block with dilation rates $r1,r2$, we convert the dilation rates to $\text{floor}(\min(r1,r2)),\text{ceil}(\max(r1,r2))$ so that the converted model maintains the same field-of-view as the original one and preserves the local details.
\subsection{Decoder}
\label{sec:decoder}

\begin{figure}
  \centering
    \includegraphics[width=0.8\linewidth]{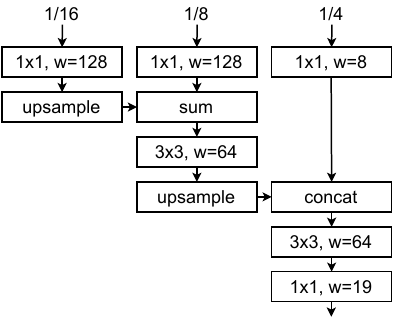}

   \caption{Decoder. $w$ shows the number of output channels. All convs except the final one are followed by BatchNorm\cite{batchnorm} and ReLU.}
   \label{fig:Decoder}
\end{figure}

The decoder's job is to restore the local details lost in the backbone. Similar to DeepLabv3+\cite{deeplabv3plus2018}, we use $[k\times k, c]$ to denote a $k\times k$ conv with $c$ output channels. We take the backbone's last $1/4$, $1/8$, and $1/16$ feature maps as inputs. We apply a $[1\times 1, 128]$ conv to $1/16$, a $[1\times 1, 128]$ conv to $1/8$ and a $[1\times 1, 8]$ conv to $1/4$. We upsample the $1/16$, sum it with the $1/8$, and apply a $[3\times 3, 64]$ conv. We upsample again, concatenate with the $1/4$, and apply a $[3\times 3, 64]$ conv, before the final $[1\times 1, 19]$ conv. All convs except the final one are followed by BatchNorm\cite{batchnorm} and ReLU. The decoder is shown in \cref{fig:Decoder}. This simple decoder performs better than many existing decoders that have similar latencies, as shown in \cref{sec:decoderComparison}.

\subsection{Dilated conv versus downsampling}
While dilated conv and downsampling are both effective ways to gain field-of-view, they both have advantages and disadvantages. Downsampling is usually computationally cheaper than dilated conv because it operates at a coarser resolution. On the other hand, dilated conv can handle thin and long objects that downsampling methods usually fail at because dilated conv operates at a high resolution and has a large field-of-view while downsampling method's coarse resolution loses representations for thin objects. Since downsampling methods have to spend more computation to upsample and retrieve the local details than dilated conv methods, they are not necessarily computationally cheaper overall. RegSeg's simple and intuitive design allows it to run in real-time while achieving competitive results.

We create a toy dataset to demonstrate RegSeg's ability to handle thin and long objects. The synthetic images are $1024\times 1024$ and contain rectangles that have the full height of $1024$ but a small width. The ground truth labels are at the bottom of the rectangles, and the model needs to propagate the labels to the top. As shown in \cref{fig:toy_comparison}, RegSeg excels at the task, only failing on extra thin rectangles of width 8 since RegSeg operates at $1/16$ of the input resolution. Downsampling methods such as DDRNet23\cite{ddrnet} and PSPNet\cite{pspnet} completely fail at the task.

\section{Experiments}

\subsection{Datasets}
Cityscapes\cite{cityscapes} is a large-scale dataset focused on street scene parsing. It contains $2975$ images for training, $500$ for validation, and $1525$ for testing. We do not use the $20000$ coarsely labeled images. There are $19$ classes and ignore label $=255$. The image size is $1024\times 2048$.

CamVid\cite{camvid} is another street scene dataset similar to Cityscapes. It contains $367$ images for training, $101$ for validation, and $233$ for testing. Following previous works\cite{dfanet,bisenetv1,swiftnet}, we use only $11$ classes and set all other classes to the ignore label $=255$. We train on the trainval set and evaluate on the test set. The image size is $720\times 960$.


\subsection{Train setting}
\label{sec:trainingSetting}
On Cityscapes, we use SGD with momentum $=0.9$, initial learning rate $=0.05$, batch size $=8$, epochs $=1000$, and weight decay $=0.0001$, but we do not decay BatchNorm parameters. We use Online Hard Example Mining loss\cite{ohem} (OHEM), which averages pixel losses that are over $0.3$ or averages the top $1/16$ pixel losses if the original proportion is less than $1/16$. We use the poly learning rate scheduler and a linear warmup\cite{warmup} from $0.1lr$ to $lr$ for the first $3000$ iterations. We apply random horizontal flipping, random scaling of $[400, 1600]$ for the shorter side while preserving the aspect ratio, and random cropping of $1024\times 1024$. We use a reduced set of RandAug\cite{cubuk2020randaugment} operations (auto contrast, equalize, rotate, color, contrast, brightness, sharpness). For each image, we apply $2$ random operations of magnitude $0.2$ (out of $1$). We also use class uniform sampling\cite{classUniformSampling} with class uniform percent $=0.5$. The weights are initialized using PyTorch's\cite{pytorch} default initialization. We use a single T4 GPU with mixed precision training. When submitting to the test server, we train on the trainval set. We also experiment with ImageNet\cite{deng2009imagenet} pretraining. In this case, we first train RegSeg for 100 epochs on ImageNet with train crop size $320\times 320$ before training 500 epochs on Cityscapes, where the training recipe on Cityscapes is the same except that we also use exponential moving average (EMA) with alpha $0.0031$ and update every 32 iterations.

On CamVid, the training setting is similar to that in Cityscapes. Since RegSeg is not pretrained on ImageNet and CamVid is small, it is initialized using Cityscapes pretrained weights. We use batch size $=12$ and epochs $=200$. We apply random horizontal flipping, random scaling of $[288, 1152]$, and random cropping of $720\times 960$. We do not apply RandAug or class uniform sampling.

\subsection{Reproducibility}

To make our ablation studies possible, we need the results to be reproducible. We sort the training images by their filenames to prevent different orders caused by different file systems. Before randomly initializing the model weights, we set the random seed to $0$. At the start of each epoch, we set the random seed to the current epoch. By doing so, we eliminate the problem of being in different states of the random number generator during training, caused by initializing different models or by resuming the model training after an incomplete training session. Furthermore, because random shuffling of the filenames happens at the start of each epoch, we can guarantee the same order of images even under different data augmentations. We run each experiment three times to get the mean and standard deviation. 

\subsection{Backbone ablation studies}
\label{sec:backboneAblation}


\begin{table}
  \centering
  \resizebox{\columnwidth}{!}{
  \begin{tabular}{l|c|c}
    \toprule
    Dilation rates & Field-of-view & mIOU\\
    \midrule
    \begin{tabular}{@{}c@{}c@{}} (1,1)+(1,2)+(1,2)+(1,3)\\+(2,3)+(2,7)+(2,3)+(2,6)\\+(2,5)+(2,9)+(2,11)+(4,7)+(5,14)\end{tabular} & 2399& $78.50\pm 0.25$\\
    \bottomrule
  \end{tabular}
  }
  \caption{The dilation rates found using differentiable neural architecture search. This method achieves comparable accuracy to the manually searched results in \cref{fig:miou_vs_fov}}
  \label{tab:dnas}
\end{table}

\begin{table}
  \centering
  \resizebox{\columnwidth}{!}{
  \begin{tabular}{l|c|c|c}
    \toprule
    backbone & context module & mIOU & FPS\\
    \midrule
    RegNetY-600MF & None & $75.66\pm 0.53$ & 31\\
    RegNetY-600MF & ASPP& $77.29 \pm 0.18$ & 28\\
    RegNetY-600MF & PPM& $77.47\pm 0.29$ & 30\\
    RegSeg & None & $\mathbf{78.65} \pm 0.28$ & 30\\
    \bottomrule
  \end{tabular}
  }
  \caption{mIOU vs backbone on Cityscapes. We fix the decoder while ablating the backbone. When using a backbone with a small field-of-view such as RegNetY-600MF, adding a context module such as PPM or ASPP increases performance. However, incorporating dilated convolution into the backbone is much more effective than adding a context module, as shown by the superiority of RegSeg. }
  \label{tab:miou_vs_backbone}
\end{table}

\begin{table}
  \centering
  \begin{tabular}{l|c}
    \toprule
    number of branches & mIOU \\
    \midrule
    2 & $78.57\pm 0.39$\\
    4 & $78.47\pm 0.56$\\
    \bottomrule
  \end{tabular}
  \caption{mIOU vs number of branches in the D block on Cityscapes.}
  \label{tab:miou_vs_number_of_branches}
\end{table}

\begin{table}
  \centering
  \begin{tabular}{l|c|c}
    \toprule
    model & mIOU & FPS\\
    \midrule
    fractional dilation rates & $78.57\pm 0.39$ & 18\\
    integral dilation rates & $78.65 \pm 0.28$& $\mathbf{30}$\\
    \bottomrule
  \end{tabular}
  \caption{Fractional dilation rates vs integral dilation rates. Rounding the dilation rates to integers gives no loss in accuracy and increases the inference speed.}
  \label{tab:fractional_vs_integral}
\end{table}
We perform many ablation studies on the backbone design. First, we experimentally confirm our claim that field-of-view is essential for the accuracy. As shown in \cref{fig:miou_vs_fov}, accuracy increases along with the field-of-view until it stabilizes when field-of-view hits around 2000. Second, we find that using 4 branches in the D block is not better than using 2 branches, as shown in \cref{tab:miou_vs_number_of_branches}. Third, we show that the model with the rounded dilation rates perform just as well as the differentiable neural searched model with the fractional dilation rates and significantly increases the inference speed, as shown in \cref{tab:fractional_vs_integral}.

\begin{figure}
  \centering
    \includegraphics[width=1.0\linewidth]{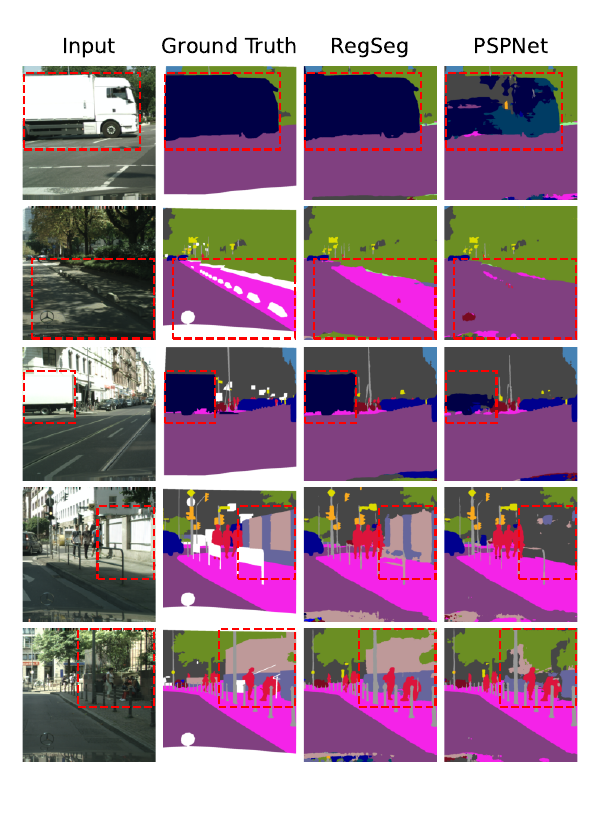}
   \caption{Comparison between RegSeg and baseline PSPNet (RegNetY600MF+PPM) on Cityscapes. RegSeg performs better than PSPNet on large objects.}
   \label{fig:cityscapes_comparison}
\end{figure}

The main motivation for the development of RegSeg is that using the dilated conv throughout the backbone should perform better than using dilated conv solely in the last layer. As shown in \cref{tab:miou_vs_backbone}, although adding a context module such as ASPP\cite{deeplabv3} or PPM\cite{pspnet} increases performance over the small-field-of-view RegNetY-600MF, RegSeg performs more than $1\%$ better than RegNetY+ASPP and RegNetY+PPM by using dilated conv throughout the backbone. \cref{fig:cityscapes_comparison} shows a qualitative comparison between RegSeg and PSPNet (RegNetY600MF+PPM).

We show the dilation rates found by the differentiable neural architecture search method in \cref{tab:dnas}. There are a few patterns that we found. The dilation rate in one branch is small to preserve the local details while the other branch uses a larger dilation rate to increase the field-of-view. The dilation rate increases over the blocks because the later blocks can use large dilation rates to the capitalize on the already increased field-of-view and because their upper bounds on the dilation rates are higher according to \cref{sec:fov}. Interestingly, the DNAS model does not always keep the smaller dilation rate in a D block to $1$ as our manual models do; instead, the small dilation rate increases to $2$, $4$, and $5$ the closer it is to the output.

\subsection{Decoder comparison}

\label{sec:decoderComparison}
\begin{table}
  \centering
  \begin{tabular}{l|c}
    \toprule
    Decoder &mIOU\\
    \midrule
    LRASPP\cite{mobilenetv32019} & $77.75 \pm 0.19$\\
    BiSeNetDecoder\cite{bisenetv1}& $77.87\pm 0.18$\\
    SFNetDecoder\cite{sfnet} &$78.12 \pm 0.38$\\
    \cref{sec:decoder} decoder& $\mathbf{78.65} \pm 0.28$\\
    \bottomrule
  \end{tabular}
  \caption{Decoder Comparison on Cityscapes. Our best decoder performs better than common alternatives.}
  \label{tab:decoderComparison}
\end{table}


In \cref{tab:decoderComparison}, we experiment with the decoder design while fixing the backbone architecture. Existing decoders\cite{bisenetv1,sfnet,mobilenetv32019} perform much worse than our best decoder, potentially because they are designed for backbones that do not have a large field-of-view. Notably, RegSeg performs around $1\%$ better than DeepLab's LRASPP.

\subsection{Timing}
\label{sec:timing}

We time RegSeg using Nvidia's T4 GPU and Mac's M1 GPU, both with mixed precision. We use PyTorch 2.0.0\cite{pytorch} with CUDA 11.8. The input size is $1\times 3\times 1024\times 2048$ on Cityscapes, and $4\times 3\times 720\times 960$ on CamVid. After $10$ iterations of warm up, we average the model's time over the next $100$ iterations. We set torch.backends.cudnn.benchmark=True and use torch.cuda.synchronize().

\subsection{Comparison on CamVid}

\begin{table}
  \centering
  \resizebox{\columnwidth}{!}{
  \begin{tabular}{l|c|c|c}
    \toprule
    Model & Extra data & mIOU & T4 FPS \\
    \midrule
    STDC2-Seg\cite{stdc} & IM & 73.9 & 152.2 \\
    GAS\cite{gas} & - & 72.8 & 153.1 \\
    CAS\cite{cas} & - & 71.2 & 169 \\
    SFNet(DF2)\cite{sfnet} & IM & 70.4 & 134 \\
    SFNet(ResNet-18)\cite{sfnet} & IM & 73.8 & 36 \\
    MSFNet\cite{msfnet} & IM & 75.4 & 91 \\
    HyperSeg-S\cite{nirkin2021hyperseg} & IM & 78.4 & 38.0\\
    TD4-PSP18\cite{td} & IM & 72.6 & 25.0 \\
    VideoGCRF\cite{VideoGCRF} & C & 75.2 & -\\
    BiSeNetV2\cite{bisenetv2} & C & 76.7 & 124 \\
    BiSeNetV2-L\cite{bisenetv2} & C & 78.5 & 33 \\
    CCNet3D\cite{huang2020ccnet} & C & 79.1 & - \\
    DDRNet-23\cite{ddrnet} & C & 80.1$\pm$0.4 & 99* \\
    \midrule
    RegSeg & C & $\mathbf{80.9}\pm0.07$ & 110*\\
    \bottomrule
  \end{tabular}
  }
  \caption{Accuracy and speed comparison on CamVid, IM: ImageNet, C: Cityscapes. DDRNet-23 and RegSeg are timed using the same environment.}
  \label{tab:camvidComparison}
\end{table}


\begin{table*}
  \centering
  \resizebox{0.9\textwidth}{!}{
  \begin{tabular}{l|c|c|c|c|c|c|c|c}
    \toprule
    Model & Extra data & val mIOU& test mIOU& T4 FPS & M1 FPS & Resolution & Params (M) & GFLOPs\\
    \midrule
    MobileNetV3\cite{mobilenetv32019} & None & 72.36 & 72.6 & 63 & 4.7 & 1024x2048 & 1.51 & 9.74\\
    BiSeNetV2-L\cite{bisenetv2} & None & 75.8 & 75.3 & 99 & 28.1 & 512x1024 & 4.59 & 139\\
    HyperSeg-M\cite{nirkin2021hyperseg} & IM & 76.2 & 75.8 & 37 & 5.9 & 512x1024 & 10.3 & 8.4\\
    FC-HarDNet-70\cite{hardnet} & None & 77.7 & 76.0 & 44 & 14.0 & 1024x2048 & 4.12 & 35.6\\
    STDC2-Seg75\cite{stdc} & IM & 77.0 & 76.8 & 41 & 17.8 & 768x1536 & 16.1 & 54.9\\
    SFNet(DF2) \cite{sfnet} & IM & - & 77.8 & 57 & 11.7 &  1024x2048 & 17.9 & 80.4\\
    DDRNet-23\cite{ddrnet} & IM &  \textcolor{black}{\textbf{79.1}} & \textcolor{black}{\textbf{79.4}} & 33  & 9.3 & 1024x2048 & 20.1 & 143.1\\
    \midrule
    RegSeg & None & 78.50 & 78.3 & 38 & 11.3 & 1024x2048 & 3.34 & 39.1\\
    RegSeg & IM & \textcolor{black}{\textbf{79.42}} & \textcolor{black}{\textbf{79.1}} & 38  & 11.3 & 1024x2048 & 3.34 & 39.1\\
    RegSeg & IM & 78.06 & 77.5 & 63 & 18.3 & 768x1536 & 3.34 & 22.0\\
    \bottomrule
  \end{tabular}
  }
  \caption{Accuracy and speed comparison on Cityscapes. IM: ImageNet. All FPS numbers are measured under our timing environment.}
  \label{tab:cityscapesComparison}
\end{table*}

As shown in \cref{tab:camvidComparison}, RegSeg achieves $80.9$ mIOU on CamVid test set at $112$ FPS. It outperforms the previous SOTA DDRNet-23 by $0.8\%$, and BiSeNetV2-L by $2.4\%$. The results show that RegSeg may generalize better than DDRNet-23. RegSeg demonstrates better data efficiency as CamVid is a small dataset.

\subsection{Comparison on Cityscapes}

\begin{table}
  \centering
  \resizebox{\columnwidth}{!}{
  \begin{tabular}{lccccc}
    \toprule
    Model & T4 FPS & M1 FPS & val mIOU & Params (M) & GFLOPs\\
    \midrule
    DDRNet-23 & 33 & 9.3 & $78.55\pm 0.35$ & 20.1 & 143.1\\
    \midrule
    RegSeg & $\mathbf{38}$ & $\mathbf{11.3}$ & $78.50 \pm 0.25$ & $\mathbf{3.34}$ & $\mathbf{39.1}$\\
    \bottomrule
  \end{tabular}
  }
  \caption{Comparison against DDRNet-23 under the exact same training setting. FPS of both models are calculated using \cref{sec:timing}. RegSeg achieves higher FPS and better parameters and flops efficiency while maintaining similar accuracy.}
  \label{tab:ddrnetComparison}
\end{table}

We compare against other real-time models on Cityscapes. As shown in \cref{fig:miouVsParams}, RegSeg achieves the best parameter-accuracy trade-offs. In \cref{tab:cityscapesComparison}, we show the accuracy and speed comparison on Cityscapes. We measure the FPS of all models in our timing environment. RegSeg outperforms HarDNet\cite{hardnet}, which is the previous SOTA model without extra data, by $1.5\%$, and outperforms SFNet(DF2)\cite{sfnet} by $0.5\%$. RegSeg also outperforms the popular BiSeNetV2-L\cite{bisenetv2} by $3.0\%$, and MobileNetV3+LRASPP\cite{mobilenetv32019} by $5.7\%$. With ImageNet pretraining, RegSeg achieves $79.4$ mIOU on the Cityscapes val set. By using a smaller resolution, $768\times 1536$, RegSeg achieves $78.06$ on the Cityscapes val set with $46$ FPS on T4 and $18.3$ FPS on M1, outperforming STDC2-Seg75\cite{stdc}. As shown in \cref{tab:ddrnetComparison}, RegSeg achieves similar results with DDRNet-23 when trained using the exact same training settings, while having better parameters, flops, and runtime efficiency. Both models are trained 5 times under our training setting described in \cref{sec:trainingSetting}.

\section{Limitations}
RegSeg's backbone might not suitable for small images such as ImageNet's usual $224\times224$ train crop size because most weights of the $3\times3$ dilated conv with large dilation rates will hit the zero padding, and the dilated conv will reduce to a normal $1\times 1$ conv.

\section{Conclusion}
In this paper, we are interested in increasing the field-of-view of the backbone by using dilated convolution with large dilation rates, while aiming for real-time performance. We demonstrate how to effectively use the dilated convolution, by having an upper bound on the dilation rate to not leave gaps in between convolutional weights and introducing the novel D block, which can increase the field-of-view in one path while preserving the local details in the other. We introduce a differentiable neural architecture search method to optimize the dilation rates by gradient descent and compare the results with the manual search metho. We also propose a lightweight decoder that performs better than common alternatives. Together, RegSeg is a radically novel approach on real-time semantic segmentation that is competitve with the state of the art on Cityscapes and CamVid datasets.

{\small
\bibliographystyle{ieee_fullname}
\bibliography{egbib}
}

\end{document}